\documentclass[conference]{IEEEtran}
\IEEEoverridecommandlockouts
\usepackage[]{footmisc}
\usepackage{cite}
\usepackage{siunitx}
\usepackage{comment}
\usepackage{lipsum}
\usepackage{amsmath,amssymb,amsfonts}
\usepackage[hidelinks]{hyperref}
\usepackage{algorithmic}
\usepackage{graphicx}
\usepackage{textcomp}
\usepackage{mathtools}
\usepackage{xcolor}
\usepackage{booktabs}
\usepackage[capitalise]{cleveref}
\crefname{figure}{Fig.}{Figs.}
\Crefname{figure}{Fig.}{Figs.}

\def\BibTeX{{\rm B\kern-.05em{\sc i\kern-.025em b}\kern-.08em
    T\kern-.1667em\lower.7ex\hbox{E}\kern-.125emX}}

\begin{document}

\title{Data-Driven or Model-Constrained? Comparing Airflow Estimation Methods under Sparse Sensing}

\author{
\IEEEauthorblockN{Dominik Aigner\IEEEauthorrefmark{1},
Thomas Wiedemann\IEEEauthorrefmark{1}\IEEEauthorrefmark{2},
Javier Monroy\IEEEauthorrefmark{3},
Achim J. Lilienthal\IEEEauthorrefmark{1}}
\IEEEauthorblockA{\IEEEauthorrefmark{1}Technical University of Munich,
\IEEEauthorrefmark{2}German Aerospace Center,
\IEEEauthorrefmark{3}University of M\'alaga\\
Email: dominik.aigner@tum.de}
}

\maketitle

\begin{abstract}
Airflow-driven advection is the primary mechanism for gas and aerosol dispersion in real-world scenarios, making accurate airflow estimation essential for ventilation monitoring, gas source localization, and contaminant spread analysis. Since obtaining airflow maps via high-resolution spatial sensing is usually infeasible in practice, flow field estimation methods must cope with sparse measurements by introducing statistical or physical priors to compensate for limited data. Existing approaches mainly differ in how they incorporate fluid-flow constraints and environmental geometry. However, due to the lack of systematic comparisons across this spectrum of methods, the potential benefits of physics-based priors for airflow estimation under sparse sensing remain underexplored.

To address this question, we quantitatively evaluate four airflow estimators across three simulated 2D scenarios under steady-state conditions and varying sampling densities. The methods range from Gaussian Process Regression and graph-based Gaussian Markov Random Fields to two Navier-Stokes-based approaches, including a novel strong-form residual minimization variant. Our results show that physical priors provide the greatest benefit at low sampling densities by compensating for missing spatial information. Although data-driven Gaussian Process Regression becomes increasingly competitive as sensor coverage increases, model-based estimators retain an accuracy advantage across the evaluated scenarios.
\end{abstract}
\begin{IEEEkeywords}
airflow estimation, physics-informed modeling, sparse sensing, inverse problems
\end{IEEEkeywords}

\section{Introduction}
The transport of airborne substances is overwhelmingly driven by advection; while molecular diffusion takes hours to spread gases or aerosols over a few meters, even subtle air currents ($< \qty{0.1}{\meter\per\second}$) achieve this in seconds \cite{nazaroff2004indoor}. Consequently, accurate airflow estimation is important for applications ranging from industrial leak monitoring to ventilation assessment in clinical settings \cite{jayaweera2020transmission, hernandez2019multi}. However, obtaining dense airflow maps through high-resolution spatial sensing is often impractical and costly, while Computational Fluid Dynamics (CFD) simulations require detailed boundary conditions and substantial computational resources. 

\begin{figure}[t]
    \centering
    \vspace*{-3pt}
    \includegraphics[width=0.9\linewidth]{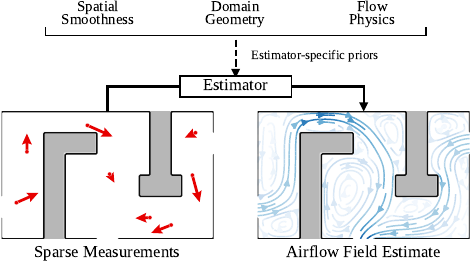}
    \caption{Illustration of a dense, steady-state airflow field estimation (right) from sparse velocity samples (left) employing different priors.}
    \label{fig:problem_illustration}
    \vspace*{-15pt}
\end{figure}

In practice, neither may be fully available, i.e., only partial boundary knowledge and sparse, noisy airflow measurements, such as those collected along a mobile robot's trajectory, are accessible. Estimating a dense airflow field under such circumstances is central to this work (see \cref{fig:problem_illustration}). We consider this problem under two simplifying assumptions: (1) we target the mean flow and therefore neglect short-term fluctuations by assuming steady state; (2) we restrict the flow field to two dimensions, as in prior work \cite{wiedemann2025physics, monroy2026physics}.

Yet, this estimation task is inherently ill-posed, as the available data underdetermine the unknown flow variables. Existing approaches tackle this challenge through regularization, ranging from statistical smoothness priors to constraints imposed by domain geometry and fluid dynamics. This raises a fundamental question: to what extent can physical priors improve airflow estimation accuracy under practically relevant sampling densities, given their increased computational complexity and the inevitable mismatch between idealized physical models and real-world flow behavior?

To assess the role of physical priors, we compare four estimators spanning this spectrum: a purely data-driven Gaussian Process Regression (GPR) \cite{williams1995gaussian}; a graph-based Gaussian Markov Random Field (GMRF)~\cite{monroy2026physics} incorporating simplified flow dynamics; and two approaches based on partial differential equations (PDEs) -- an existing weak-form Navier-Stokes Minimization (WFNS)~\cite{wiedemann2025physics} and a novel strong-form Navier-Stokes Residual Minimization (SFNS) variant introduced in \cref{sec:nsrm_method}. In summary, our contributions are twofold: (i) the novel SFNS estimator, and (ii) a systematic comparison across this spectrum using OpenFOAM CFD simulations under varying sampling densities (\Cref{sec:comparison}).

\section{Strong-form Navier-Stokes Residual Minimization}
\label{sec:nsrm_method}
The proposed strong-form Navier-Stokes Residual Minimization (SFNS), a variant of~\cite{wiedemann2025physics}, combines sparse velocity measurements with a stationary physical flow model. The available measurements are denoted by $\{(x_i,z_i)\}_{i=1}^{K}$, where $x_i \in \Omega$ is the measurement location and $z_i \in \mathbb{R}^2$ is the corresponding measured velocity. The objective is to estimate a steady-state velocity field $u:\Omega\rightarrow\mathbb{R}^2$, with the kinematic pressure field $p:\Omega\rightarrow\mathbb{R}$ treated as an auxiliary variable. This leads to the optimization problem:
\begin{equation} \label{eq:min_NS}
    (\hat{u}, \hat{p})
    =
    \arg\min_{u,p}
    \underbrace{
    \left(
    E_m + \lambda_{\mathrm{NS}}E_{\mathrm{NS}} + \lambda_s E_s
    \right)
    }_{\coloneqq J(u,p)} .
\end{equation}
The measurement term $E_m = \sum_{i=1}^{K}\|u(x_i)-z_i\|^2$ penalizes deviations from the sensor measurements, while ${E_s=\int_{\Omega}\|\nabla u\|^2\,d\Omega}$ adds a smoothness prior with a weight factor $\lambda_s$. The Navier-Stokes term $E_{\mathrm{NS}}$, weighted by $\lambda_{\mathrm{NS}}$, penalizes violations of the momentum and incompressibility equations with kinematic viscosity $\nu$:
\begin{equation}\label{eq:E_NS}
    E_{\mathrm{NS}}
    =
    \int_{\Omega}
    \left( \|(u\cdot\nabla)u + \nabla p -\nu \Delta u\|^2
    + (\nabla \cdot u)^2
    \right)
    d\Omega .
\end{equation}
To avoid solving a nonlinear problem in \eqref{eq:min_NS}, we linearize the convective term in \eqref{eq:E_NS} in a fixed-point iteration:
\begin{equation}
    (u\cdot\nabla)u
    \approx
    (u_{k-1}\cdot\nabla)u_k .
\end{equation}
At each iteration $k = 1,2,\ldots$, the advection velocity is taken from the previous step, starting from $(u_0,p_0)=0$~\cite{rehman2008numerical}. This turns $J(u,p)$ into a sequence of quadratic functionals $J_k(u_k,p_k)$ with linear first-order optimality conditions $\delta J_k = 0$. Despite slower convergence, this so-called Picard scheme does not require additional derivatives and is more robust for higher-Reynolds-number flows compared to Newton linearization~\cite{schwarz2018data}. 

To obtain a finite-dimensional problem, $u_k$ and $p_k$ are approximated on a finite-element (FE) mesh using Taylor-Hood elements, i.e., piecewise quadratic basis functions for velocity and piecewise linear basis functions for pressure \cite{elman2014finite, taylor2024finite}. With $N_u$ velocity nodes and $N_p$ pressure nodes, the nodal coefficients are collected in the global state vector
\begin{equation}
    \mathbf{w}_k =
    (\mathbf{u}_k,\mathbf{p}_k)^{\top}
    \in \mathbb{R}^{M},
    \qquad
    M \coloneqq 2N_u + N_p,
\end{equation}
where $\mathbf{u}_k \in \mathbb{R}^{2N_u}$ stacks both velocity components at all $N_u$ nodes and $\mathbf{p}_k \in \mathbb{R}^{N_p}$ contains the pressure coefficients.

Applying the FE approximation to the linear normal equations derived from $\delta J_k = 0$ yields a linear system per iteration:
\begin{equation}
    \left(
    \mathbf{S}
    + \mathbf{K}(\mathbf{u}_{k-1})
    + \mathbf{R}
    \right)
    \mathbf{w}_k
    =
    \mathbf{S}\mathbf{w}_{\mathrm{meas}} .
\end{equation}
Here, $\mathbf{K}(\mathbf{u}_{k-1}) \in \mathbb{R}^{M\times M}$ denotes the weighted contribution of the discretized and linearized Navier-Stokes residual, and $\mathbf{R} \in \mathbb{R}^{M\times M}$ is the weighted contribution of the smoothness regularization. To align coordinates, each sensor position $x_i$ is mapped to its nearest mesh node. The matrix $\mathbf{S} \in \mathbb{R}^{M\times M}$ is then a binary diagonal selection matrix where $S_{ii}=1$ if the $i$-th degree of freedom (DOF) corresponds to a velocity component of a measured node, and zero otherwise. The vector $\mathbf{w}_{\mathrm{meas}}$ contains the measured velocities $z_i$ at these active DOFs and zeros elsewhere.

After assembly, no-slip Dirichlet conditions are imposed on DOFs associated with walls and obstacles. The pressure is fixed at one open boundary \cite{taylor2024finite}. Solving the resulting linear systems over the Picard iterations updates $\mathbf{w}_k$ toward a dense estimate that balances measurement consistency, strong-form flow residual minimization, and smoothness regularization.

\section{Experimental Comparison}
\label{sec:comparison}
Following the presentation of SFNS, we return to the central question posed in this paper: evaluating the influence of physical priors on the estimation of 2D airflow fields. We therefore compare our proposed approach against the three baselines introduced above: GPR, GMRF, and WFNS.

\subsection{Experimental Setup}
We evaluated the methods in three simulated indoor scenarios. Their spatial domains were discretized into rectangular occupancy grids for GPR and GMRF, and into triangular meshes for the FE-based SFNS and WFNS. To support a fair comparison, we chose discretizations with a comparable number of DOFs. Physics-informed methods additionally had access to partial boundary information, distinguishing between walls and open boundaries, but inflow and outflow were assumed to be unknown. For each scenario, the precomputed CFD solution $u^*$ served as both the reference field for accuracy assessment and the source of sparse measurements. Measurements were drawn at $K=10,\ldots,250$ random locations to decouple the estimator comparison from the potential influence of a particular sampling strategy. They were also corrupted with zero-mean Gaussian noise ($\sigma=\qty{0.1}{\meter\per\second}$) in both velocity components, a typical uncertainty level for robotic anemometers \cite{wiedemann2025physics}. We evaluated each estimate $\hat{u}$ using the mean absolute magnitude error $e_{\mathrm{ME}}$ and the mean angular error $e_{\mathrm{AE}}$ over $N_c$ free-space points:
\begin{align}
    e_{\mathrm{ME}}
    &=
    \frac{1}{N_c}
    \sum_{j=1}^{N_c}
    \left|
    \|\hat{u}(x_j)\|
    -
    \|u^*(x_j)\|
    \right|, \\
    e_{\mathrm{AE}}
    &=
    \frac{1}{N_c}
    \sum_{j=1}^{N_c}
    \arccos
    \left(
    \frac{
    \hat{u}(x_j) \cdot u^*(x_j)
    }{
    \|\hat{u}(x_j)\|
    \|u^*(x_j)\|
    }
    \right).
\end{align}
Angular errors exclude areas where flow direction becomes ill-defined, i.e., with ground-truth speeds below the noise standard deviation $\sigma$. To assess computational cost, we additionally report the estimator runtime. All metrics are averaged over three $10\times\qty{6}{\meter}$ indoor scenarios with different obstacle layouts.

All methods used fixed hyperparameters across the evaluation. GMRF and WFNS used the values recommended in their respective publications \cite{monroy2026physics, wiedemann2025physics}, while we manually tuned the length scale of GPR to $\qty{0.25}{\meter}$. For SFNS, we set $\{\lambda_{\mathrm{NS}},\lambda_s\}=\{10,0.01\}$ and $\nu=\qty{1.5e-5}{\meter\squared\per\second}$ to approximate air. The Picard iteration was terminated after 25 steps or when $\|\mathbf{w}_k-\mathbf{w}_{k-1}\|/\|\mathbf{w}_k\|<0.01$.

\begin{figure}
  \centering
  \includegraphics[width=\linewidth]{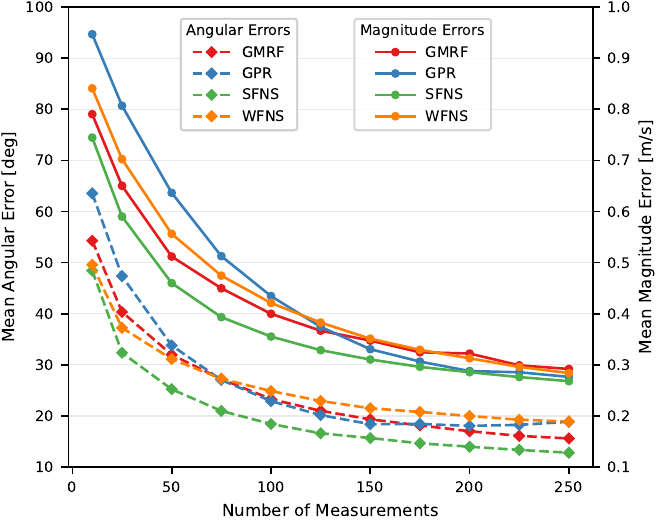}
  \caption{Combined estimation errors for increasing numbers of measurements. Solid lines denote magnitude error, dashed lines denote angular error. All results are averaged over 10 random sample sets and three different scenarios; uncertainty bands are omitted for readability.}
  \label{fig:results_plot}
  \vspace*{-10pt}
\end{figure}

\subsection{Results and Discussion}
Across all methods, the curves for $e_{\mathrm{ME}}$ and $e_{\mathrm{AE}}$ show the expected asymptotic behavior with increasing numbers of available measurements (see \cref{fig:results_plot}), decreasing sharply in the low-sample regime before gradually flattening. 

SFNS achieves the lowest overall velocity magnitude errors. It outperforms the other methods at low sampling densities; however, this gap narrows as more measurements are available. In particular, GPR becomes increasingly competitive as the sample size grows, but does not outperform SFNS. At $K=250$, all methods reach similar average magnitude errors slightly below $\qty{0.3}{\meter\per\second}$. 

For flow direction, SFNS consistently achieves the lowest angular errors. GPR improves strongly with additional measurements but remains less accurate than SFNS, while GMRF and WFNS also remain less accurate across all settings.

These results support the hypothesis that physical priors compensate for sparse measurement data: the performance gap between data-driven and model-based estimators is largest in the low-sample regime and narrows as more measurements become available. Nevertheless, GPR does not surpass SFNS in the dense-sampling regime. This advantage should be interpreted cautiously, as CFD-based reference fields may provide a favorable setting for the PDE-based estimators. A plausible explanation for GPR's remaining gap is the fixed length scale, chosen as a compromise across all sampling densities and therefore not optimized for each setting.

However, this accuracy advantage comes with computational cost (see \cref{tab:runtime}). GPR is by far the fastest, followed by GMRF and the proposed SFNS, while WFNS requires approximately one second per run. Notably, all evaluated methods are computationally suitable for online robotic applications, where the physical collection of spatial samples typically consumes the majority of the time budget. 

\begin{table}[b]
    \centering
    \vspace*{-10pt}
    \renewcommand{\arraystretch}{1.3}
    \caption{Mean Estimator Runtimes in Milliseconds}
    \begin{tabular*}{\columnwidth}{@{\extracolsep{\fill}}ccccc@{}}
        \hline
        $K$ & GPR & GMRF & SFNS & WFNS \\
        \hline
        50  & $1.9\pm1.1$ & $67.9\pm37.7$ & $341.6\pm66.0$ & $1030.9\pm285.1$ \\
        100 & $3.6\pm2.2$ & $60.4\pm18.7$ & $331.6\pm83.5$ & $975.0\pm246.9$ \\
        200 & $8.6\pm5.1$ & $51.8\pm14.8$ & $323.0\pm98.9$ & $903.9\pm251.1$ \\
    \end{tabular*}
    \label{tab:runtime}
\end{table}

\begin{figure}
    \centering
    \includegraphics[width=1.0\linewidth]{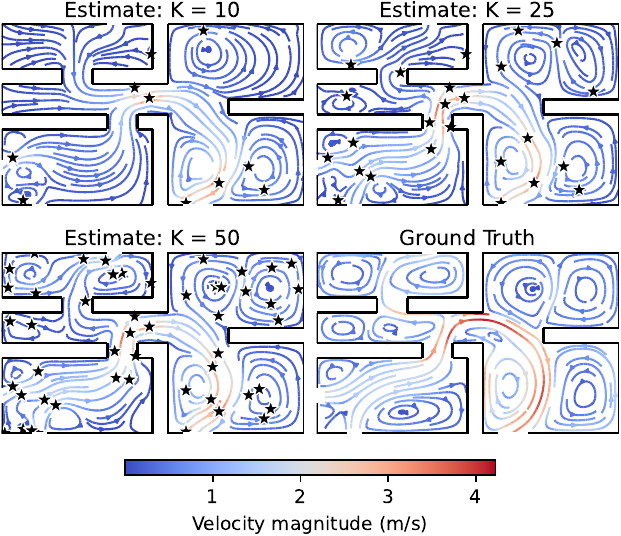}
    \caption{SFNS estimates for low sampling densities, with ground truth shown for comparison. Stars indicate random measurement locations. The example shows that even low sampling can recover the main flow-field structure.}
    \label{fig:estimation_increasing_samples}
    \vspace*{-13pt}
\end{figure}

For SFNS, this corresponds to a threefold speedup over the WFNS predecessor, indicating that directly penalizing the strong-form residual not only improves accuracy but can also be computationally attractive. \Cref{fig:estimation_increasing_samples} shows how SFNS manages to represent the main characteristics of the reference flow field even for low sampling densities. Still, GMRF's competitive accuracy relative to WFNS -- combined with very short execution times -- demonstrates that complex physical priors alone do not guarantee better results.

For some applications, e.g., gas source localization, directional accuracy is crucial. According to \cite{wiedemann2019analysis}, the average angular wind error should remain below $\qty{30}{\degree}$ for reasonable source localization. In our scenarios, SFNS falls below this threshold from about 50 random measurements onward. Using a targeted flow-field exploration strategy, even fewer measurements may be sufficient.

\section{Conclusions and Future Work}
In this work, we compared representative methods for estimating 2D airflow fields. In addition to existing data-driven, graph-based, and weak-form Navier-Stokes approaches, we introduced a strong-form Navier-Stokes Residual Minimization variant. In simulation-based evaluations using noisy measurements sampled from CFD reference fields, the proposed SFNS performed particularly well in scenarios with sparse measurement data, which are prevalent in real-world applications. Although data-driven methods such as Gaussian Process Regression became increasingly competitive as more measurements became available, SFNS retained an advantage, particularly in terms of directional accuracy.

Potential future work includes extending this comparison to real-world experiments and to more complex, or even three-dimensional, domains. From a methodological perspective, automating the currently manual hyperparameter selection would improve the practical applicability of model-based estimators and reduce potential tuning bias.


\bibliographystyle{ieeetr}
\newpage
\bibliography{literature}
\end{document}